%% file: RISeg2.tex
\documentclass[letterpaper, 10 pt, conference]{ieeeconf}
\IEEEoverridecommandlockouts
\usepackage{cite}
\usepackage{amsmath,amssymb,amsfonts}
\usepackage{algorithm}
\usepackage[noend]{algpseudocode}
\usepackage{graphicx}
\usepackage{textcomp}
\usepackage{xcolor}
\usepackage{multirow}
\usepackage{siunitx}
\usepackage{adjustbox}
\usepackage{array}
\usepackage{tikz}
\usepackage{pgfplots}
\usepackage{caption}
\usepackage{hyperref}
\usepackage{dsfont}
\usepackage{makecell}
\usepackage{xcolor}
\usepackage[normalem]{ulem}

\pgfplotsset{compat=1.18}
\usepackage[font=small]{caption}

\pgfplotsset{
every axis/.append style={
  axis line style={->}, 
  legend style={font=\footnotesize},
  label style={font=\footnotesize},
  title style={font=\footnotesize},
  tick label style={font=\footnotesize},
  axis x line*=bottom,
  axis y line*=left,
  }
}

\def\BibTeX{{\rm B\kern-.05em{\sc i\kern-.025em b}\kern-.08em
    T\kern-.1667em\lower.7ex\hbox{E}\kern-.125emX}}
\begin{document}
\title{\bf \LARGE rt-RISeg: Real-Time Model-Free Robot Interactive Segmentation\\for Active Instance-Level Object Understanding}

\author{Howard H. Qian, Yiting Chen, Gaotian Wang, Podshara Chanrungmaneekul, Kaiyu Hang
\thanks{The authors are with the Department of Computer Science, Rice University, Houston, TX 77005, USA. This project is supported by the US National Science Foundation grant FRR-2133110.}
}

\maketitle

\begin{abstract}
Successful execution of dexterous robotic manipulation tasks in new environments, such as grasping, depends on the ability to proficiently segment unseen objects from the background and other objects. Previous works in unseen object instance segmentation (UOIS) train models on large-scale datasets, which often leads to overfitting on static visual features. This dependency results in poor generalization performance when confronted with out-of-distribution scenarios. To address this limitation, we rethink the task of UOIS based on the principle that vision is inherently interactive and occurs over time. We propose a novel real-time interactive perception framework, rt-RISeg, that continuously segments unseen objects by robot interactions and analysis of a designed body frame-invariant feature (BFIF). We demonstrate that the relative rotational and linear velocities of randomly sampled body frames, resulting from selected robot interactions, can be used to identify objects without any learned segmentation model. This fully self-contained segmentation pipeline generates and updates object segmentation masks throughout each robot interaction without the need to wait for an action to finish. We showcase the effectiveness of our proposed interactive perception method by achieving an average object segmentation accuracy rate 27.5\% greater than state-of-the-art UOIS methods. Furthermore, although rt-RISeg is a standalone framework, we show that the autonomously generated segmentation masks can be used as prompts to vision foundation models for significantly improved performance.
\end{abstract}

\begin{figure}[t]
\includegraphics[width=0.99\columnwidth]{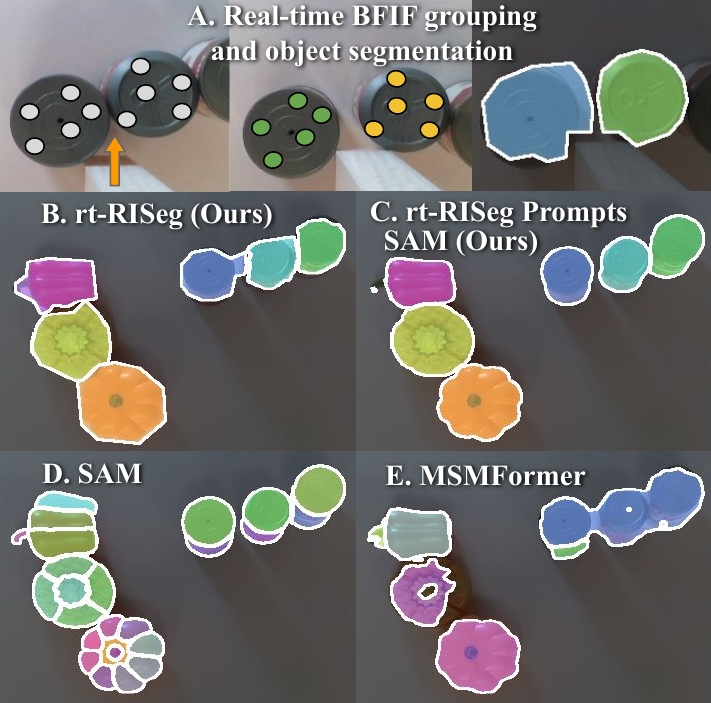}
\centering
\captionsetup{skip=2pt}
\caption{Interactive segmentation of a cluttered scene through minimally disruptive actions with comparisons to state-of-the-art learned static segmentation models. [A] Real-time BFIF grouping and object segmentation throughout a robot interaction. The orange arrow indicates the robot's action direction. Frames are sampled initially and grouped once scene motions are detected. Only a small amount of rigid body motion is needed for accurate segmentation, which preserves the initial task formation. Object segmentation is achieved in real time. [B] rt-RISeg segmentation resulting from robot interactions and [C] rt-RISeg autonomously prompting Segment Anything {\color{black}Model}. Static segmentation models [D] Segment Anything {\color{black}Model (SAM)} and [E] {\color{black}Mean Shift Mask Transformer (MSMFormer)}.}
\label{mainFig}
\vspace{-18pt}
\end{figure}

\section{Introduction}

A robot's ability to segment unseen objects in new environments is essential to scene understanding and successful execution of downstream tasks, such as dexterous manipulation. Therefore, robust unseen object instance segmentation (UOIS) is critical for any robotic system aiming to perform real-world tasks~\cite{RISeg,Back2022Amodal,Xie2020UOIS,Xiang2021UOIS,Danielczuk2018}.

State-of-the-art UOIS methods train computationally expensive deep neural networks by using large datasets to extract latent space feature representations. However, under- and over-segmentation remains a challenge for these learned models in out-of-distribution (OOD) scenarios, such as heavily cluttered environments~\cite{Xiang2021UOIS, MSMFormer}. For these methods, segmentation is performed on static images and is only considered a pixel grouping problem in the 2-dimensional image space. However, objects physically exist in the 3-dimensional world and have unique properties when under manipulation, such as relative rigid-body motions. Interactive perception utilizes this insight and approaches UOIS through robot manipulation over continuous time rather than at a single static moment~\cite{Brock2009}. These robot-object interactions should aim to be minimally disruptive to avoid breaking the initial task formation of the scene~\cite{RISeg}. For example, if the robot's task is to place a coffee mug into a cabinet, initial interactive segmentation of the coffee mug should be achieved with as little physical disruption as possible to avoid accidentally knocking over and breaking the mug.

Central to rt-RISeg is the designed body frame-invariant feature (BFIF)~\cite{RISeg}, which uses the insight that two body frames rigidly attached to the same moving object will have the same spatial twist observed from any fixed world frame, no matter their actual motions~\cite{ModernRobotics}. Extending this principle to multiple objects allows for robust object differentiation, even with slight differences in movement. Since this feature is free of learning, there is minimal lag time between rigid body motions and object identification. Moreover, unlike previous frameworks that approach interactive perception through ``observe, interact, observe", we support a shift in paradigm towards ``observe while interacting"~\cite{RISeg, MARTINEZContPercept}. In addition to more accurate object segmentation, this shift, along with the BFIF, provides an opportunity for immediate downstream task planning, since objects can be segmented in real time throughout robot interactions. 

We build on this core BFIF insight and propose rt-RISeg, a lightweight, model-free interactive perception framework that robustly segments objects in real time while maintaining initial task formation through minimally disruptive interactions (see Fig.~\ref{mainFig}). rt-RISeg leverages BFIFs to derive object-level understanding without any learning, and BFIF-based object segmentation occurs concurrently with robot scene interactions. Furthermore, the resulting instance-level object understanding can be used to prompt vision foundation models, such as Segment Anything {\color{black}Model} (SAM)~\cite{kirillov2023segmentanything}, to generate near perfect refined unseen object instance segmentation masks (as shown in Fig.~\ref{mainFig}-C).

\begin{figure}[t]
\vspace{5pt}
\includegraphics[width=0.85\columnwidth]{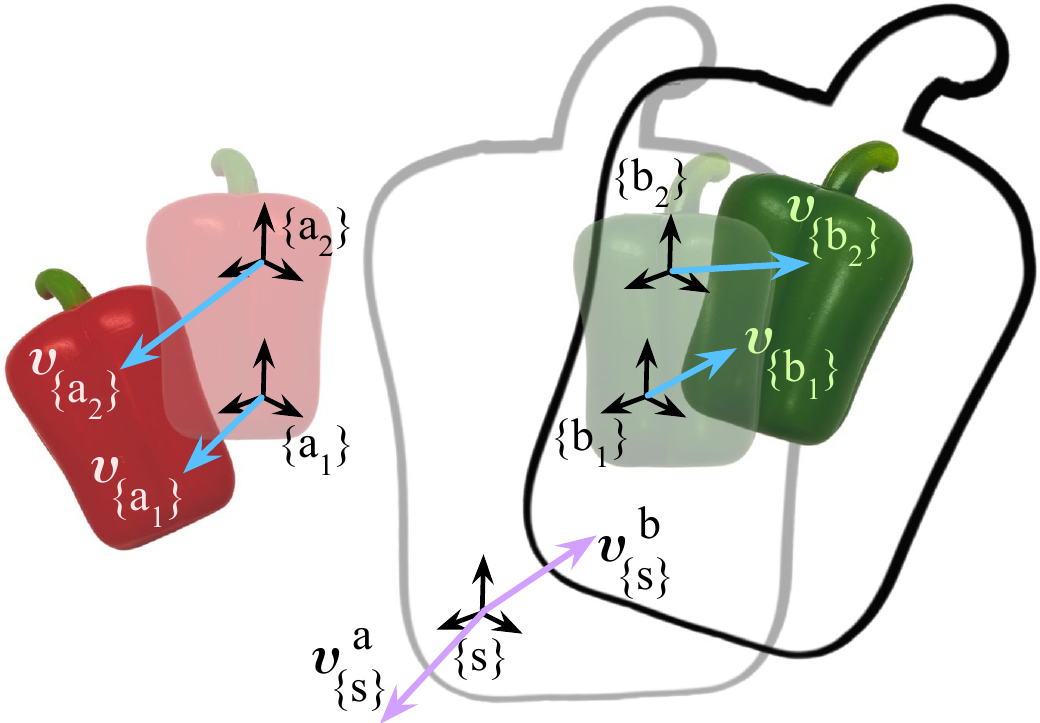}
\centering
\captionsetup{skip=2pt}
\caption{A visual representation of BFIFs, where the different motions of body frames attached to the same rigid body are the same when transformed into a fixed space frame. Transparent objects represent the initial positions, and solid objects represent the subsequent positions due to some motions. Sampled body frames $\{a_1\}$, $\{a_2\}$, $\{b_1\}$ and $\{b_2\}$ lie on the initial configurations of their respective objects and space frame $\{s\}$ is arbitrarily chosen. The translation of body frames resulting from the rigid body motions are represented as the blue linear velocity vectors $\upsilon_{x}$. Although $\upsilon_{\{a_1\}} \neq \upsilon_{\{a_2\}}$ and $\upsilon_{\{b_1\}} \neq \upsilon_{\{b_2\}}$, the transformation to the fixed space frame results in the same linear velocity vector $\upsilon_{\{s\}}^x$ for each body frame on the same rigid object $x$ but different linear velocity vectors for different rigid objects. The outlined pepper shape imagines the green object to be infinitely large, which shows that different motions of different body frames transformed into one fixed space frame will always be equal.}
\label{SpatialTwist}
\vspace{-18pt}
\end{figure}

\section{Related Work}
\subsection{Unseen Object Instance Segmentation}

Unseen object instance segmentation is the task of segmenting every object instance within a scene without any pre-existing object-specific knowledge~\cite{Xiang2021UOIS}. In early UOIS work, low-level image features, such as edges, convexity, and contours, were examined for object identification~\cite{Felzenszwalb2004Graph, Pham2017SceneCut, Stein2014Convexity, forsyth2003computer, Trevor2013EfficientOP}. However, because these methods lack an object-level understanding and analyze each characteristic of an image, scenes are often over-segmented. Recently, deep neural networks have been trained on large-scale datasets to produce learned models that drastically improve segmentation performance~\cite{Xie2019Both, Danielczuk2019Data, Shao2018ClusterNetIS, kirillov2023segmentanything, MSMFormer,Xiang2021UOIS}. Despite these improvements, such computationally intensive methods still struggle to generalize to OOD scenarios, often resulting in segmentation inaccuracy due to the challenges of bridging the sim-to-real gap and dataset biases~\cite{Zhang2023UOIS,Balloch2018Synthetic,kirillov2023segmentanything}. Since both low-level and learning-based methods propose to segment objects from static images, their performance is limited by a lack of real-world object-level understanding. Therefore, we propose a model-free interactive perception framework that drastically outperforms these state-of-the-art UOIS methods through autonomous robot interactions and analysis of rigid body motions for physical understanding.

\subsection{Motion-Based Object Perception}
Motion-based segmentation methods approach object segmentation from an active lens, using scene changes to segment objects~\cite{Brock2009, Bohg2016}. Early work utilized statistical and factorization methods, which either required prior knowledge or were computationally expensive~\cite{zappella2008motion, Costeira1998AMF, Goh2007, Arsenio, metta2003early}. Multi-view segmentation methods leverage multiple camera angles to capture images and discern consistencies for more accurate object detection~\cite{Zeng2017MultiView,Mitash2017MultiView}. However, like static segmentation models, these methods struggle with challenges associated with a lack of intuition about the physical rigid body properties of objects in the real world~\cite{Brock2009}. The proposed rt-RISeg method addresses these challenges by introducing a lightweight framework that analyzes small object motions for object-level real-world understanding and segmentation.

\subsection{Interactive Perception}
Interactive perception for UOIS is the practice where a
robot physically interacts with its environment
to segment objects~\cite{RISeg}. Recent work used robot actions to singulate objects for accurate segmentation and back propagation~\cite{lu2023selfsupervised}. However, this approach breaks the initial task formation and is not viable for many meaningful dexterous manipulation tasks where objects may be fragile or the general environment state needs to be preserved. RISeg was introduced to correct segmentation inaccuracies of existing UOIS models~\cite{RISeg}. However, this framework requires base segmentation and uncertainty masks from a learned model to analyze pairs of images captured before and after each interaction. Our proposed rt-RISeg is a model-free, real-time interactive perception framework that continuously segments objects through minimally disruptive interactions, preserving task formation. Since the lightweight framework does not utilize a learned model for base segmentation masks, we demonstrate that rigid body motions of interacted objects can be directly used to autonomously segment unseen objects and produce object-level understanding without external guidance.

\section{Preliminaries and Problem Formulation}
In this section, we will first define the designed body frame-invariant feature central to the rt-RISeg object identification and segmentation pipeline. Then, we will formally define the interactive perception problem and introduce our proposed method.

\subsection{Body Frame-Invariant Feature}
The proposed rt-RISeg interactive perception framework utilizes a designed body frame-invariant feature (BFIF) to analyze rigid body motions and differentiate objects from one another for UOIS. In RISeg, we proposed this body frame-invariant feature (BFIF), which is inspired by spatial twists of body frames attached to moving rigid bodies. The central idea is that the twists of different moving body frames attached to the same rigid body expressed in a fixed space frame will have the same spatial twist, regardless of their absolute motions~\cite{ModernRobotics}.

Given a rigid body that experiences some motion in space, the translation and rotation of an attached body frame $\{b\}$ can be represented as a twist $\mathcal{V}_b$, which is defined as

\begin{align}
\label{twistNotation}
\mathcal{V}_b = [\omega_b, \upsilon_b]^\intercal \in \mathbb{R}^6
\end{align}
where angular velocity vector $\omega_b$ and linear velocity vector $\upsilon_b$ are represented in the $\{b\}$ frame. However, since the instantaneous body twists of multiple body frames attached to the same moving rigid body will be different, we transform each body twist into a spatial twist $\mathcal{V}_s = [\omega_s, \upsilon_s]^\intercal \in \mathbb{R}^6$, represented in a common fixed space frame \{s\}. 

For moving body frame \{b\} and fixed space frame \{s\}, let $T_{sb}$ be the transformation matrix from \{s\} to \{b\} and $\dot T_{sb}$ be the time derivative of $T_{sb}$. For body twist $\mathcal{V}_b$, the equivalent spatial twist $\mathcal{V}_s$ is derived by
\begin{align}
\label{twistDerivation}
\dot T_{sb}T_{sb}^{-1} = \begin{bmatrix} [\omega_s] & \upsilon_s \\ 0 & 0 \\ \end{bmatrix}  = [\mathcal{V}_s]
\end{align}
where $[\omega_s]_{3\times 3}$ is the skew-symmetric representation of $\omega_s$. Therefore, spatial twist can be calculated for each body frame.

Fig.~\ref{SpatialTwist} illustrates the intuition behind spatial twist that if we imagine a rigid body to be infinitely large, the instantaneous linear velocity $\upsilon_s$ (and angular velocity $\omega_s$) of the space frame \{s\} will be equal for all body frames \{b\} on the same rigid body~\cite{ModernRobotics}. The \emph{Body Frame-Invariant Feature} (BFIF), also denoted by $\mathcal{V}_s$, represents the concept of spatial twist.

\subsection{Problem Formulation}
The interactive perception approach to UOIS attempts to segment unseen objects in an environment through autonomous robot interactions over time. As previously mentioned, rt-RISeg proposes a paradigm shift in interactive perception from ``observe, interact, observe" to ``observe while interacting".

To formalize this, let the observation data $I_t \in [0, 255]^{H \times W \times 3} \times \mathbb{R}_{+}^{H \times W}$ be an RGB-D image of the given scene at time step $t$, where $t = 0, 1, 2, \ldots$, is the discrete time of the system. Furthermore, let $\theta_t \in \mathbb{R}^n$ be the joint angles at time step $t$ of a robot with $n$ joints. We aim to segment each scene by observing object motions throughout robot interactions. Let $a_i \in SE(3)$ be a robot action, where $i$ is the discrete interaction step of the system. With each new image $I_t$ and robot joint angles $\theta_t$ produced during the execution of $a_i$, we produce the segmentation mask $L_{t} \in \mathbb{Z}_{+}^{H \times W}$, where the object ID given by $L_t^{(i,j)} \in L_t$ corresponds to the pixel $(i,j)$ in the image $I_t$. A value $L_t^{(i,j)} = 0$ segments the pixel $(i,j)$ in the image $I_t$ as part of the background. Any integer value $L_t^{(i,j)} > 0$ indicates that the pixel $(i,j)$ belongs to an object with the unique object ID $L_t^{(i,j)}$. 

In Alg.~\ref{rt-RISeg}, we algorithmically describe the rt-RISeg system in which objects are actively segmented while the robot interacts with the environment. After a robot action $a_i$ is identified by $\Call{FindAction}{\cdot}$, the segmentation mask $L_{t}$ is continuously updated via BFIF analysis so long as action $a_i$ is still being executed by $\Call{Interact}{\cdot}$, as shown in lines 5-8. After action $a_i$ is completed, $L_t$ is stored in $L_t^*$ as an intermediate result used only for evaluation purposes. Once the stop condition $a_i = \textit{null}$ is met, the segmentation mask $L_t$ is returned, which represents the final object segmentation of the scene's end configuration after all interactions.

\begin{figure}[t]
\vspace{-5pt}
\begin{algorithm}[H]
\footnotesize
\caption{rt-RISeg}\label{rt-RISeg}
\textbf{Input:} \textit{none} \\
\textbf{Output:} segmentation mask, $L_{t}$
\begin{algorithmic}[1]
\State $t, i \leftarrow 0, 0$
\State $\theta_t, I_t \leftarrow \Call{GetCurrRobot}{t}, \Call{GetCurrImage}{t}$
\State $L_{t} \leftarrow \mathbf{0}_{H \times W}$
\While {$a_i \leftarrow \Call{FindAction}{L_{t}, I_t}$ \textbf{not} \textit{null}} \Comment{Alg.~\ref{FindAction}}
\While {$\Call{Interact}{a_i}$} \Comment{real-time segmentation}
\State $t \leftarrow t+1$
\State $\theta_t, I_t \leftarrow \Call{GetCurrRobot}{t}, \Call{GetCurrImage}{t}$
\State $L_{t} \leftarrow \Call{SegmentObjs}{I_{t-1}, I_{t}, \theta_{t-1}, \theta_{t}, {L}_{t-1}}$ \Comment{Alg.~\ref{SegmentObjs}}
\EndWhile
\State $i \leftarrow i+1$
\State $L_{t}^* \leftarrow L_{t}$ \Comment{*intermediate result*}
\EndWhile
\State \textbf{return} $L_{t}$
\end{algorithmic}
\end{algorithm}
\vspace{-30pt}
\end{figure}

\section{Robot Interactive Object Segmentation}

In this section, we will first describe how robot actions are heuristically selected using the segmentation mask and depth image at the current time step. Then, we will describe how objects are segmented throughout each interaction via optical flow-based BFIF grouping. 

\subsection{Action Selection}
In Alg.~\ref{FindAction}, we detail an action selection algorithm with two goals. First, we select actions such that moved objects do not share the same motion, which ensures effective object differentiation through BFIF analysis. Second, we select actions that minimally disrupt the environment so as to not jeopardize the scene's initial task formation. We assume that unseen objects lie on a flat tabletop.

In line 1 of Alg.~\ref{FindAction}, given the current RGB-D image of the scene $I_t$, $\Call{GetObjsAboveTable}{\cdot}$ uses the depth channel of $I_t$ and performs random sample consensus (RANSAC){\color{black}~\cite{fischler1981ransac}} for plane fitting to generate a binary mask $objMask \in \{0, 1\}^{H \times W}$, where $objMask^{(i,j)} = 1$ indicates that pixel $(i,j)$ belongs to an object on the tabletop, while $objMask^{(i,j)} = 0$ indicates that it does not. Then, the current system's segmentation mask $L_{t}$ is binarized and defined as $binL \in \{0, 1\}^{H \times W}$, where
\begin{align}
\label{binarizeDefinition}
binL^{(i,j)} = \begin{cases} 
1, & \text{$L_{t}^{(i,j)} > 0$} \\
0, & \text{$L_{t}^{(i,j)} = 0$}
\end{cases} 
\end{align}
for all $(i, j)$. We subtract $objMask$ and $binL$ to produce a binary mask $objsToSegment \in \{0, 1\}^{H \times W}$ representing the objects that are yet to be segmented. If binary mask $objsToSegment$ is entirely filled with zeros, then the scene is considered fully segmented, and a \textit{null} action is returned.

\begin{figure}[t]
\vspace{-5pt}
\begin{algorithm}[H]
\footnotesize
\caption{FindAction}\label{FindAction}
\textbf{Input:} $L_{t}, I_t$ \\
\textbf{Output:} $a_i$
\begin{algorithmic}[1]
\State $objMask \leftarrow \Call{GetObjsAboveTable}{I_t}$
\State $binL \leftarrow \Call{BinarizeMask}{L_{t}}$
\State $objsToSegment \leftarrow objMask - binL$
\If{$objsToSegment == \mathbf{0}_{H \times W}$}
\State \textbf{return} \textit{null}
\EndIf
\State $centers, clusterPoints \leftarrow \Call{KMeans}{objsToSegment}$
\State $boundaryPoints \leftarrow \Call{Boundary}{clusterPoints}$
\For{$(c, P, B)$ in zip($centers, clusterPoints, boundaryPoints$)}
\For{$b$ in $B$}
\If{\Call{IsValidPush}{$b, \overrightarrow{bc}, d_{push}, \ell_{act}, P, objMask$}}
\State $a_i \leftarrow (b, \overrightarrow{bc}, d_{push})$
\State \textbf{return} $a_i$
\EndIf
\EndFor
\EndFor
\State \textbf{return} \textit{null}
\end{algorithmic}
\end{algorithm}
\vspace{-30pt}
\end{figure}

In line 6 of Alg.~\ref{FindAction}, the $objsToSegment$ binary mask is used to identify cluster centers $centers$ and corresponding cluster points $clusterPoints$. These clusters are computed via K-Means{\color{black}~\cite{macqueen1967kmeans}} on pixels $(i,j) \in objsToSegment \;\forall\; objsToSegment^{(i,j)} > 0$, with $k$ chosen by the elbow method. {\color{black}While the limitations of K-Means and the elbow method are acknowledged, the proposed action selection algorithm is robust to variations in clustering and operates under minimal requirements, as detailed in the next paragraph. Alternative clustering techniques, such as K-Medoids~\cite{kmedoids} or Mean Shift~\cite{meanshift}, could also be used effectively.} Then, for each cluster center and its corresponding points, the boundary points $boundaryPoints$ are derived by $\Call{Boundary}{\cdot}$. Each cluster center and boundary point pair is a candidate for robot action selection.

\begin{figure}[t]
\vspace{5pt}
\includegraphics[width=0.99\columnwidth]{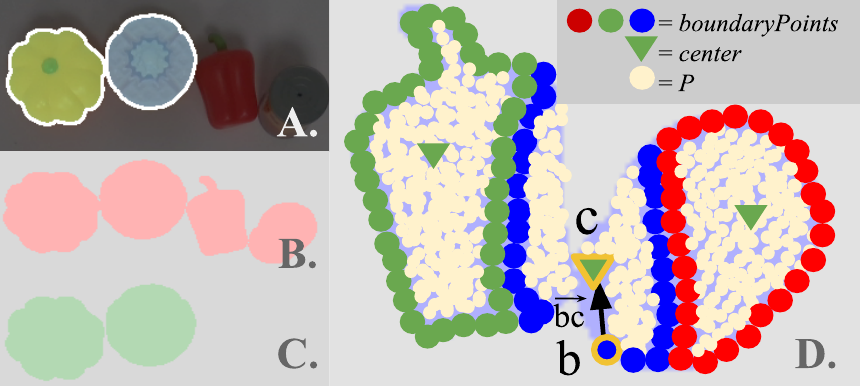}
\centering
\captionsetup{skip=2pt}
\caption{Alg.~\ref{FindAction} FindAction($\cdot$) heuristically selecting a robot action $a_i$. [A] Current system segmentation mask $L_t$ visualized over input image $I_t$. [B] $objMask$ represents pixels belonging to objects. [C] $binL$ represents pixels already segmented. [D] $objsToSegment$ represents pixels belonging to objects that still need to be segmented with $boundaryPoints$, $center$, and $P$ overlaid. IsValidPush($\cdot$) evaluates a candidate push starting at boundary point $b$ in direction $\overrightarrow{bc}$ towards center $c$. Candidate action is selected if $b$ is accessible to the robot and if the translation of cluster points $P$ in the direction $\overrightarrow{bc}$ over a distance $d_{push}$ does not overlap with many other known object pixels in $objMask$.}
\label{findActionFig}
\vspace{-18pt}
\end{figure}

In lines 8 through 12 of Alg.~\ref{FindAction}, for each cluster center $c$ and a corresponding boundary point $b$, we return a robot action $a_i$ if the action starting at $b$ in the direction of $\overrightarrow{bc}$ for push distance $d_{push}$ is valid. Shown in Fig.~\ref{findActionFig}, a ``valid" push is defined by $\Call{IsValidPush}{\cdot}$ through {\color{black}3 criteria on the cluster center,} contact point, and push direction. {\color{black}First, cluster center $c$ must lie on an object, denoted by $objMask^{c} = 1$.} Second, push point $b$ must be accessible by the robot. Formally, $b$ meets the second criteria if an area in the direction $\overrightarrow{cb}$ a short distance from $b$ is free of obstacles, determined by $objMask^{(i,j)} = 0$ for all $(i,j)$ in the area needed for the robot end effector. Third, the push direction $\overrightarrow{bc}$ must not cause the moved object to immediately move other objects in the same manner. This ensures differences in spatial twists when BFIF analysis is performed for object differentiation. To validate this, the set of cluster points $P$ is translated a distance of $d_{push}$ in the direction $\overrightarrow{bc}$. The push direction $\overrightarrow{bc}$ is valid if the translated cluster does not have a large overlap with other points in $objMask$. Formally, we calculate the intersection $isect$ of translated points $P'$ and $objMask$ to be
\begin{align}
\label{validPTranslation}
isect = P' \cap objMask \quad\text{ where } objMask^P = 0
\end{align}
If $|isect|/|P| \leq \ell_{act}$, where $\ell_{act}$ is a constant threshold, then the robot push direction $\overrightarrow{bc}$ is valid, and action $a_i$ is returned. If no valid push exists, then a \textit{null} action is returned.

\subsection{Real-Time Object Segmentation}
In Alg.~\ref{SegmentObjs}, we introduce a real-time object segmentation algorithm that analyzes rigid body motions resulting from robot-object interactions. The main components of this algorithm are illustrated in Fig.~\ref{segObjs}, which include BFIF sampling, BFIF grouping, and object segmentation.

\subsubsection{Body Frame Sampling}
Lines 1 through 4 in Alg.~\ref{SegmentObjs} describe the main components in sampling body frames from moving pixels. The proposed rt-RISeg framework attaches a camera to the robot's wrist to reduce real-time occlusion of objects throughout interactions. Though not necessary for a general real-time interactive perception framework, this choice in implementation results in a moving camera, which means that the motion must be subtracted from the observed optical flow to understand which points experienced effective motion in the real world as opposed to motion relative to the camera. First, the RGB-D image $I_{t-1}$ and consecutive robot joint angles $\theta_{t-1}$ and $\theta_t$ are used to calculate $E_t \in \mathbb{R}^{H \times W \times 2}$, the expected flow due to the camera's motion. This is done by converting the depth data from $I_{t-1}$ to 3D coordinates $dMap_{t-1} \in \mathbb{R}^{H \times W \times 3}$ represented in the camera frame and deriving the camera transformation matrix $camT_{t-1,t}$ from $\theta_{t-1}$ and $\theta_t$ {\color{black}via forward kinematics~\cite{kucuk2006forwardkinematics}}. The expected 3D coordinates at $t$ is then defined as
\begin{align}
\label{dMapPrime}
dMap_t = camT_{t-1,t}
\begin{bmatrix}
dMap_{t-1} \\
1
\end{bmatrix}
\end{align}
where $dMap_t$ is represented in homogeneous coordinates. {\color{black}$dMap_t$ is then projected back to 2D pixel coordinates $proj_t \in \mathbb{R}^{H \times W \times 2}$, where each element $proj_t^{(i,j)}$ represents the expected 2D pixel coordinate at time step $t$ for the point originally located at pixel $(i,j)$ at time step $t-1$, based solely on camera motion. The expected flow $E_t$ is then calculated by subtracting the original pixel coordinates from $proj_t$.}
\begin{align}
\label{projt}
{\color{black}E_t^{(i,j)} = proj_t^{(i,j)} - (i,j) \quad\forall (i,j)}
\end{align}

After the observed optical flow $O_t \in \mathbb{R}^{H \times W \times 2}$ is produced by $\Call{OpticalFlow}{I_{t-1}, I_t}$, the effective optical flow of objects in the real world $X_t \in \mathbb{R}^{H \times W \times 2}$ is derived from element-wise subtraction $O_t-E_t$. If there is no effective motion caused by robot action $a_i$ between $t-1$ and $t$, then $\|X_t^{(i,j)}\|_2 \approx 0 \;\forall\; (i,j)$, where $\|X_t^{(i,j)}\|_2$ is the Euclidean norm of the effective optical flow at pixel $(i,j)$. 

To compute and analyze BFIFs for object segmentation, we create body frames $\{F_{t-1}^k\}$ attached to rigid bodies in $t-1$ and track their motions through to $t$. Because $X_t$ represents absolute motion in the real world as a result of a robot interaction, we know that pixels of interest are $(i,j) \in X_t$ where $\|X_t^{(i,j)}\|_2 >> 0$. Thus, we sample $n$ random pixels from $\|X_t^{(i,j)}\|_2 >> 0$ and pick triplets of pixels to create each frame. Each triplet must not be collinear and must have a maximum pairwise distance $d_a$. A point in the triplet is chosen to be the origin, while the other two points are used to find directions for each axis. The \textit{x}-axis is perpendicular to the plane formed by the triplet of sampled points, the \textit{y}-axis is formed by connecting the origin with one of the other two points, and the \textit{z}-axis is perpendicular to the \textit{x} and \textit{y} axes. These sampled frames $\{F_{t-1}^k\}$ are then tracked through to $t$ using $X_t$ to create $\{F_{t}^k\}$.

\begin{figure}[t]
\vspace{-8pt}
\begin{algorithm}[H]
\footnotesize
\caption{SegmentObjs}\label{SegmentObjs}
\textbf{Input:} $I_{t-1}, I_{t}, \theta_{t-1}, \theta_t, {L}_{t-1}$ \\
\textbf{Output:} ${L}_{t}$
\begin{algorithmic}[1]
\State $E_t \leftarrow \Call{GetExpectedFlow}{I_{t-1}, \theta_{t-1}, \theta_t}$
\State $O_{t} \leftarrow \Call{OpticalFlow}{I_{t-1}, I_{t}}$
\State $X_t \leftarrow O_t-E_t$ \Comment{subtract camera motion}
\State $\{F_{t-1}^k\}, \{F_{t}^k\} \leftarrow \Call{CreateFrames}{X_t, \theta_t}$
\State $\{\mathcal{V}_{t}^k\} \leftarrow \Call{CalcBFIFs}{\{F_{t-1}^k\}, \{F_{t}^k\}}$
\State $G_{t} \leftarrow \Call{GroupBFIFs}{\{\mathcal{V}_{t}^k\}}$
\State $L_{t} \leftarrow \Call{Segment}{G_t, O_t, L_{t-1}}$
\State \textbf{return} $L_{t}$
\end{algorithmic}
\end{algorithm}
\vspace{-33pt}
\end{figure}

\subsubsection{BFIF Computing and Grouping}
A set of body frame-invariant features $\{\mathcal{V}_{t}^k\}$ represented in a space frame $\{s\}$ can be calculated using corresponding sets of body frames $\{F_{t-1}^k\}$ and $\{F_{t}^k\}$. {\color{black}I}n line 5 of Alg.~\ref{SegmentObjs}{\color{black}, $\Call{CalcBFIFs}{\cdot}$ calculates }each BFIF $\mathcal{V} \in \{\mathcal{V}_{t}^k\}$ by deriving the transformation matrices $T_{sb}$ and time derivative $\dot T_{sb}$ between each body frame pair ($F_{t-1}^{k}, F_{t}^{k}$) and the space frame $\{s\}${\color{black}, as described by Equation~\ref{twistDerivation}}. In this work, $\{s\}$ is selected to be the camera frame at $t-1$ for simplicity. While BFIFs $\mathcal{V} \in \{\mathcal{V}_{t}^k\}$ are theoretically equal if they are attached to the same rigid body, noise in effective optical flow $X_t$ causes slight inaccuracies, which need to be statistically filtered out for accurate BFIF grouping and object differentiation.

\begin{figure}[t]
\vspace{3pt}
\includegraphics[width=0.99\columnwidth]{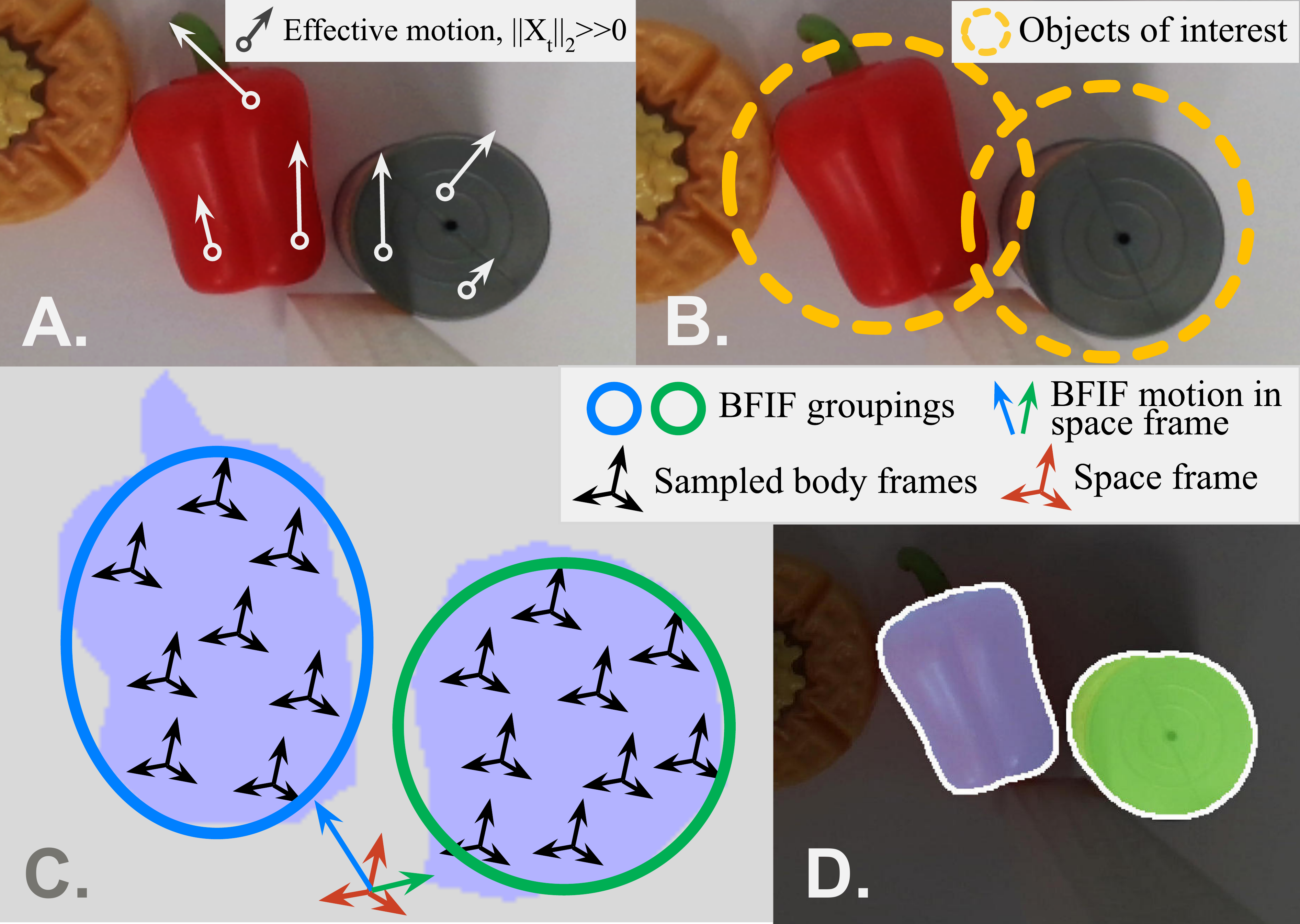}
\centering
\captionsetup{skip=2pt}
\caption{Alg.~\ref{SegmentObjs} SegmentObjs($\cdot$) uses images ($I_{t-1}$, $I_t$) and robot joint angles ($\theta_{t-1}$, $\theta_t$) to derive effective optical flow $X_t$. Body frames $\{F_{t-1}^k\}$ and $ \{F_{t}^k\}$ are sampled at points where $\|X_t\|_2>>0$. BFIFs are then computed and analyzed for real-time segmentation mask seeding. [A] Effective motion resulting from minimally disruptive robot action $a_i$. Note that effective motion at each point on each moving object is different, meaning BFIF grouping is needed for object identification. [B] Objects of interest derived from $\|X_t\|_2>>0$. [C] Body frames are sampled from pixels $(i,j) \in \|X_t\|_2>>0$ . Analysis of BFIFs results in object-level grouping. [D] Each body frame in a group is used as a seed point to flood fill an accurate segmentation mask. Pictured is real-time object segmentation by rt-RISeg prompts SAM.}
\label{segObjs}
\vspace{-17pt}
\end{figure}

\begin{figure*}[t]
\vspace{3pt}
\includegraphics[width=2\columnwidth]{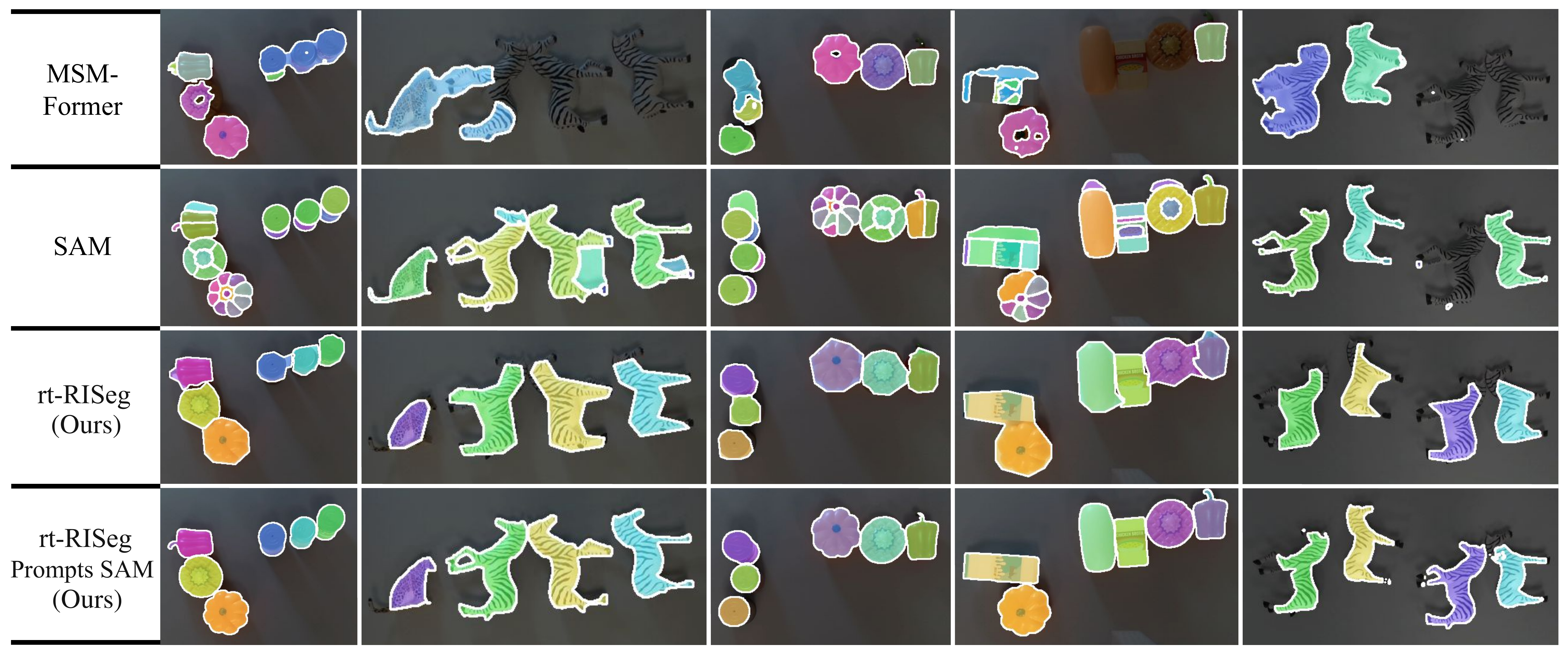}
\centering
\captionsetup{skip=2pt}
\caption{Qualitative comparison of unseen object instance segmentation between MSMFormer~\cite{MSMFormer}, SAM~\cite{kirillov2023segmentanything}, rt-RISeg (Ours), and rt-RISeg Prompts SAM (Ours). Each column showcases a different scene after robot interactions are completed. Each row showcases a method's object segmentation. MSMFormer consistently under-segments each scene, while SAM consistently over-segments each scene. rt-RISeg effectively segments objects and autonomously prompting SAM increases accuracy.}
\label{QualitativeResults}
\vspace{-13pt}
\end{figure*}

In line 6 of Alg.~\ref{SegmentObjs}, BFIFs $\{\mathcal{V}_{t}^k\}$ are grouped for object differentiation by computing pairwise BFIF Mahalanobis distances~\cite{Mahalanobis} and applying Markov clustering~\cite{van2000graph} (shown in Fig.~\ref{segObjs}-C). Mahalanobis distance is used to compare the multi-dimensional twist vectors of sampled frames by incorporating their covariance structure, making it well-suited for measuring similarity given the feature space. The resulting distances are transformed into similarity values by a Gaussian kernel and define a weighted graph where nodes represent the BFIFs of sampled frames and edge weights represent their statistical similarity. Markov clustering is then applied. This unsupervised graph clustering method simulates random walks by iteratively reinforcing strong connections and weakening weak ones through expansion and inflation operations. Upon convergence, BFIF groups $G_t$ are formed, where each group $g \in G_t$ corresponds to a connected component in the clustered graph.

\subsubsection{Object Segmentation and Forward Propagation}
In line 7 of Alg.~\ref{SegmentObjs}, $\Call{Segment}{\cdot}$ uses the grouped BFIFs $G_t$, optical flow $O_t$, and the previous segmentation mask $L_{t-1}$ to propagate previous masks to the current frame and segment new objects resulting from motion between $t-1$ and $t$.

First, accumulations of previous segmentation masks from all robot interaction steps $i$ and system time steps $t$ are propagated to the current frame $t$ by using $L_{t-1}$ and observed optical flow $O_t$. Formally, propagation of previous segmentations is defined as $L_{t}^{(x,y)} = L_{t-1}^{(i,j)}  \;\forall\; L_{t-1}^{(i,j)} > 0, \; \text{where } (x,y) = (i,j) + (a,b) \in O(t)^{(i,j)}$.

Then, new object segmentations are added to $L_t$ using the BFIF groupings $G_t$. Remember that each BFIF is associated with a body frame attached to a rigid body in our scene. If there were no physical robot-to-object interactions between $t-1$ and $t$, then $X_t \approx {0}^{H\times W}$ and body frame sets $F_{t-1}$ and $F_t$ would be empty. In this case, $L_t$ is returned because BFIF groupings $G_t$ would also be empty. Otherwise, each set $g \in G_t$ contains body frames identified to have equal BFIFs. Thus, each body frame in $g$ will be segmented as one object in mask $L_t$. First, the corresponding pixel of each body frame in $g$ is set to an object ID $\ell \notin L_t$, meaning a new object is detected and segmented. These initial assignments act as seed points for object $\ell$ since the number of sampled pixels $n$ used for body frame construction in line 4 of Alg.~\ref{SegmentObjs} is very small relative to the number of pixels $H\times W$. Afterwards, Breadth First Search starts from each seed point and is used to assign object ID $\ell$ to pixels that move with similar gradient as to neighboring pixels already assigned to $\ell$.

Rather than simply using optical flow gradients, BFIFs must be used for the initial seeding of newly segmented objects because BFIFs are body frame-invariant (shown in Fig.~\ref{segObjs}). Once each set $g \in G_t$ is segmented, the real-time segmentation mask $L_t$ is returned.

\section{Experiments}
In this section, we will showcase that the proposed rt-RISeg framework is effective for real-time interactive unseen object segmentation in cluttered environments by comparison with state-of-the-art methods {\color{black}Mean Shift Mask Transformer (MSMFormer)}~\cite{MSMFormer} and Segment Anything {\color{black}Model} (SAM)~\cite{kirillov2023segmentanything}. Our experiments demonstrate that unseen object segmentation can be interactively performed without any learned segmentation model and without disrupting the initial task formation. Although rt-RISeg can run as a lightweight standalone method, we exhibit near perfect object segmentation mask boundaries when using rt-RISeg masks to autonomously prompt vision foundation models.

\subsection{Implementation and Experiment Setup}
\subsubsection{rt-RISeg implementation} The rt-RISeg interactive perception framework uses a Franka Emika Panda robot~\cite{FrankaEmika} to interact with scene objects. An Intel Realsense D415 RGB-D Camera~\cite{D415} is used to capture real-time visual data and is rigidly attached to the wrist of the robot end-effector, always facing down in line with the vertical \textit{z}-axis. A 30cm long 3D printed rectangular end-effector extension provides adequate field-of-view throughout interactions. In line 10 of Alg.~\ref{FindAction}, constants $d_{push}$ and $\ell_{act}$ were introduced to heuristically validate a candidate action. $d_{push}$ indicates the distance of each robot action $a_i$ and is defined as 2cm. $\ell_{act}$ is the threshold for the maximum allowed intersection area divided by cluster area as a result of moving the candidate cluster and is defined as $0.3$. In line 2 of Alg.~\ref{SegmentObjs}, Recurrent All-Pairs Field Transforms for Optical Flow (RAFT)~\cite{RAFT} is used, though any off-the-shelf optical flow method can be substituted without affecting the framework's generality.

\begin{figure}[t]
\vspace{-3.5pt}
\includegraphics[width=0.95\columnwidth]{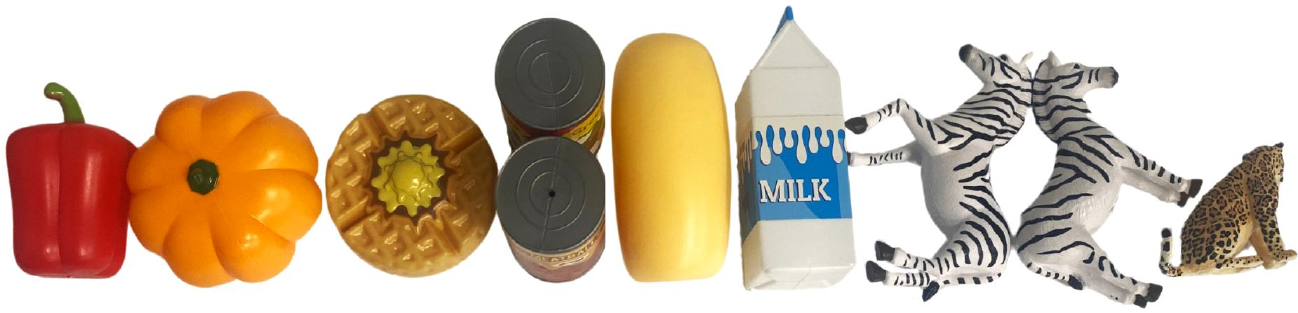}
\centering
\captionsetup{skip=2pt}
\caption{Experiment objects chosen for segmentation difficulty.}
\label{expObjs}
\vspace{-15pt}
\end{figure}

\subsubsection{Experiment setup and dataset} 
Since no standard interactive perception dataset exists, we evaluate our pipeline by creating 20 scenes in which 4-6 objects are placed in close proximity to one another and lie on a flat, white tabletop. Experiment objects come from a set of toy food items and patterned animals shown in Fig.~\ref{expObjs}, which are particularly difficult to segment due to shape and color. For each scene, 3-5 robot interactions are autonomously identified and completed for real-time segmentation, resulting in the evaluation of roughly 100 total images. Ground truth masks are manually annotated for each of these images. 

\subsection{Evaluation Metrics}
\begin{figure}[t]
\vspace{-1pt}
\resizebox{0.47\textwidth}{!}{\input{RISeg2_figures/SSRfig}}
\centering
\captionsetup{skip=2pt}
\caption{Percentage of objects correctly segmented by each method across scene configurations resulting from robot actions, measured by the Overlap F-measure $\geq 75\%$ and averaged across all experiment scenes. Although rt-RISeg is able to consistently segment about 90\% of objects after all interactions, based on our observations, the 75.1\% object segmentation accuracy suggests that rt-RISeg object boundary refinement is not very accurate. Using rt-RISeg to prompt SAM greatly improves boundary segmentation but does not dramatically improve true object overlap segmentation due to the area of segmentation being very similar.}
\label{SSRgraph}
\vspace{-15pt}
\end{figure}
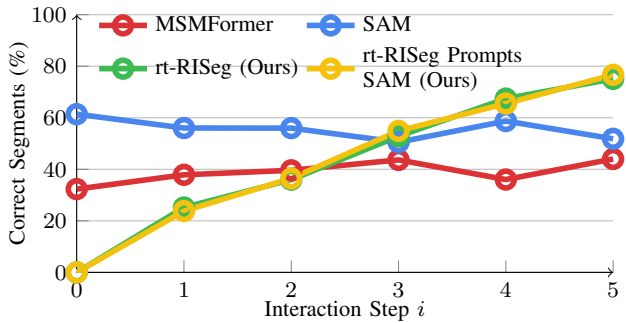

For each experiment scene, we evaluated the accuracy of rt-RISeg segmentation masks ${L}_t^*$, which are the segmentation masks produced at the end of each robot interaction $a_i$, where interaction time step $i=0$ is the initial scene configuration, and $i>0$ is the scene configuration after interaction $a_i$ is completed. In Table~\ref{MetricsTable}, Fig.~\ref{QualitativeResults}, and Fig.~\ref{SSRgraph}, we quantitatively and qualitatively compare segmentation results of our rt-RISeg method to state-of-the-art UOIS models MSMFormer~\cite{MSMFormer} and SAM~\cite{kirillov2023segmentanything}. We also showcase refined segmentation masks when using rt-RISeg to autonomously prompt SAM. For Table~\ref{MetricsTable} and Fig.~\ref{SSRgraph}, the evaluation statistics at $i$ indicate averages across all experiment scenes after the interaction $a_i$ is complete. 

Object segmentation performance is evaluated using precision, recall, and F-measure on both Overlap and Boundary criteria~\cite{Xiang2021UOIS, Xie2019Both}. These evaluation metrics for rt-RISeg are shown in Table~\ref{MetricsTable}, along with the same metrics for MSMFormer and SAM. In addition to precision, recall, and F-measure, Fig.~\ref{SSRgraph} shows the percentage of objects in a scene segmented with a high accuracy at each configuration. High accuracy is defined as segmentation of an object with Overlap F-measure $\geq 75\%$.

\subsection{rt-RISeg Prompting Foundation Models}
As mentioned previously, rt-RISeg is a standalone interactive perception framework that autonomously segments unseen objects in real time and without a learned segmentation model. Yet, some learned segmentation models offer impressive performance in boundary segmentation accuracy. A few of these models are also able to be prompted for segmentation of specific regions in a given image. Segment Anything {\color{black}Model} is one with strong boundary recognition but frequently over-segments objects when unsupervised. In Fig.~\ref{QualitativeResults}, we qualitatively showcase how segmentation masks autonomously generated by rt-RISeg can be used to prompt SAM for near perfect boundaries. Additionally, Table~\ref{MetricsTable} and Fig.~\ref{SSRgraph} quantitatively showcase the improved segmentation accuracy when rt-RISeg is used to prompt SAM.

\subsection{Discussion of Results}
Interaction step $i=0$ indicates the scene's initial configuration, in which rt-RISeg does not segment any objects. Shown in Fig.~\ref{SSRgraph}, rt-RISeg incrementally segments objects in the scene with each robot interaction, eventually surpassing static segmentation accuracy after 2-3 robot interactions. After all interactions, rt-RISeg is able to accurately segment 75.1\% of objects in a given scene, measured by Overlap F-measure $\geq 75\%$, while static segmentation models only accurately segment between 45\% and 50\%. Because rt-RISeg is able to consistently produce segmentation masks for about 90\% of objects in a given scene, based on our observations, this 75.1\% metric indicates that rt-RISeg is not very accurate in refining object mask boundaries, which is supported by Boundary statistics in Table~\ref{MetricsTable}.

In addition to an increasing trend in object segmentation accuracy, the rt-RISeg Overlap and Boundary P/R/F metrics also increase with each robot interaction, surpassing the learned segmentation models, MSMFormer and SAM, after 2-3 robot interactions. On average, rt-RISeg outperforms learned models in object segmentation accuracy by about 27.5\%, Overlap F-measure by about 18.5\%, and Boundary F-measure by about 12.5\%, after all interactions. Furthermore, using rt-RISeg to autonomously prompt SAM only slightly improves object-level segmentation accuracy, but drastically improves Boundary P/R/F metrics. This is because the segmented area remains similar to the rt-RISeg segmentation masks, but the boundaries are much more accurate.

\begin{table}[t]
\vspace{5pt}
\centering
\renewcommand{\arraystretch}{0.871}
\LARGE
\small
\captionsetup{labelformat=empty} 
\begin{adjustbox}{width=8.65cm,center}
\setlength{\tabcolsep}{5pt}
\begin{tabular}{|c|c|c c c|c c c|}           
\hline 
\multirow{2}{*}{Method} &\multirow{2}{*}{Step $i$}    & \multicolumn{3}{c|}{Overlap}  & \multicolumn{3}{c|}{Boundary}                   \\   
 &  &{P} & {R} & {F}   &{P} & {R} & {F}           \\   \hline 
\multirow{6}{*}{MSMFormer~\cite{MSMFormer}} 
 & 0 & 76.9 & 40.7 & 45.7 & 55.1 & 31.7 & 31.6                \\
 & 1 & 79.9 & 43.5 & 48.9 & 56.8 & 34.8 & 34.4                   \\   
 & 2 & 80.7 & 42.1 & 49.6 & 53.1 & 33.9 & 35.0                    \\ 
 & 3 & 84.0 & 44.5 & 51.2 & 61.7 & 37.0 & 38.7                    \\ 
 & 4 & 77.9 & 40.9 & 48.2 & 54.2 & 35.3 & 36.3                    \\ 
 & 5 & 77.0 & 51.0 & 59.0 & 47.9 & 40.3 & 41.5              \\   \hline
 \multirow{6}{*}{SAM~\cite{kirillov2023segmentanything}} 
 & 0 & 63.6 & 59.0 & 61.1 & 35.3 & 62.1 & 43.7                \\
 & 1 & 62.7 & 53.3 & 55.2 & 36.3 & 54.3 & 39.3                  \\   
 & 2 & 61.7 & 55.2 & 57.8 & 34.9 & 57.8 & 41.5                    \\   
 & 3 & 58.6 & 52.8 & 55.0 & 33.3 & 55.3 & 39.4                    \\ 
 & 4 & 60.0 & 57.5 & 58.6 & 31.6 & 58.5 & 39.9                    \\ 
 & 5 & 57.0 & 52.3 & 54.1 & 29.3 & 53.5 & 35.9              \\   \hline
 \multirow{6}{*}{\makecell{rt-RISeg\\(Ours)}}
 & 0 & 100.0 & 0.0 & 0.0 & 100.0 & 0.0 & 0.0                \\
 & 1 & 89.2 & 19.6 & 30.5 & 54.8	& 12.5 & 19.2                   \\   
 & 2 & 92.7 & 34.5 & 48.4 & 54.2	& 21.5 & 30.0                    \\ 
 & 3 & 92.9 & 48.3 & 60.6 & 51.7	& 30.0 & 37.0                    \\ 
 & 4 & 89.9 & 59.9 & 70.1 & 57.1	& 39.8 & 46.3                    \\   
 & \textbf{5} & \textbf{85.2} & \textbf{71.8} & \textbf{75.4} & \textbf{55.8} & \textbf{49.3} & \textbf{51.8}              \\   \hline
 \multirow{6}{*}{\makecell{rt-RISeg\\prompts SAM\\(Ours)}}
 & 0 & 100.0 & 0.0 & 0.0 & 100.0 & 0.0 & 0.0                   \\
 & 1 & 98.4 & 21.1 & 32.4 & 86.7 & 21.3 & 31.8     \\     
 & 2 & 98.4 & 36.5 & 51.6 & 80.0 & 34.4 & 46.8                    \\ 
 & 3 & 98.5 & 51.6 & 65.2 & 80.1 & 49.7 & 59.5                    \\ 
 & 4 & 96.8 & 61.9 & 74.1 & 82.6 & 61.2 & 69.5                \\    
 & \textbf{5} & \textbf{91.4} & \textbf{73.4} & \textbf{79.5} & \textbf{79.1} & \textbf{72.4} & \textbf{74.9}                \\   
 \hline
 \end{tabular}
 \end{adjustbox}
 \captionsetup{skip=2pt}
 \caption{\label{MetricsTable}Table 1: Segmentation results of MSMFormer, SAM, rt-RISeg and rt-RISeg prompts SAM across scene configurations resulting from robot actions. rt-RISeg outperforms learned models after all interactions in Overlap and Boundary P/R/F metrics. Using rt-RISeg to prompt SAM slightly improves Overlap metrics and drastically improves Boundary metrics.}
 \vspace{-15pt}
 \end{table}

\section{Conclusion}
In this paper, we proposed a real-time model-free interactive perception framework, rt-RISeg, which uses non-disruptive interactions and Body Frame-Invariant Feature (BFIF) analysis for accurate unseen object segmentation. The designed feature uses the insight that two body frames attached to the same moving rigid body will have the same spatial twist observed by any fixed world frame, even if their rotations and translations in space are different. Without any learned segmentation model, rt-RISeg uses the BFIF for fully self-contained object segmentation. We then demonstrated the effectiveness of rt-RISeg in autonomously segmenting real-world cluttered tabletop scenes without breaking each scene's initial task formation. Finally, we showcased how rt-RISeg can be used to prompt learned methods in an unsupervised manner for highly accurate segmentation masks. {\color{black}A potential limitation in this work is presented by the wrist-mounted camera, as it requires balancing a tradeoff between maximizing field of view and minimizing robot-object occlusion. Future work could explore incorporating a next-best-view algorithm through the use of a secondary robot arm dedicated to perception, which would complement the interactive exploration conducted by the primary arm.}

\bibliographystyle{IEEEtran}
\bibliography{RISeg2}

\end{document}

%% file: RISeg2_figures/SSRfig.tex
%
%
\definecolor{mycolor1}{rgb}{0.85882,0.19608,0.21176}%
\definecolor{mycolor2}{rgb}{0.28235,0.52157,0.92941}%
\definecolor{mycolor3}{rgb}{0.23529,0.72941,0.32941}%
\definecolor{mycolor4}{rgb}{0.95686,0.76078,0.05098}%
\begin{tikzpicture}

\begin{axis}[%
width=0.78\columnwidth,
height=0.375\columnwidth,
scale only axis,
xmin=0,
xmax=5,
xlabel style={at={(axis description cs:0.5,-0.07)}, anchor=north},
xlabel={Interaction Step $i$},
xtick={0,1,...,5},
ymin=0,
ymax=100,
ylabel style={at={(axis description cs:-0.07,0.5)}, anchor=south},
ylabel={Correct Segments (\%)},
axis background/.style={fill=white},
title style={font=\bfseries},
ymajorgrids,
legend style={at={(0.02,0.85)}, anchor=west, legend cell align=left, align=left, draw=white!15!black, fill=none, draw=none, legend columns = 2}
]
\addplot [color=mycolor1, line width=2.0pt, mark size=3.0pt, mark=o, mark options={solid, mycolor1}]
  table[row sep=crcr]{%
0	32.315\\
1	37.870\\
2	39.608\\
3	43.646\\
4	36.026\\
5	44.000\\
};
\addlegendentry{MSMFormer}

\addplot [color=mycolor2, line width=2.0pt, mark size=3.0pt, mark=o, mark options={solid, mycolor2}]
  table[row sep=crcr]{%
0	61.389\\
1	56.019\\
2	55.980\\
3	50.521\\
4	58.718\\
5	51.833\\
};
\addlegendentry{SAM}

\addplot [color=mycolor3, line width=2.0pt, mark size=3.0pt, mark=o, mark options={solid, mycolor3}]
  table[row sep=crcr]{%
0	0.000\\
1	25.185\\
2	35.980\\
3	52.813\\
4	67.436\\
5	75.100\\
};
\addlegendentry{rt-RISeg (Ours)}

\addplot [color=mycolor4, line width=2.0pt, mark size=3.0pt, mark=o, mark options={solid, mycolor4}]
  table[row sep=crcr]{%
0	0.000\\
1	23.796\\
2	36.471\\
3	54.896\\
4	65.513\\
5	76.667\\
};
\addlegendentry{rt-RISeg Prompts\\SAM (Ours)}

\end{axis}
\end{tikzpicture}%

%% file: RISeg2.bbl
\begin{thebibliography}{10}
\providecommand{\url}[1]{#1}
\csname url@samestyle\endcsname
\providecommand{\newblock}{\relax}
\providecommand{\bibinfo}[2]{#2}
\providecommand{\BIBentrySTDinterwordspacing}{\spaceskip=0pt\relax}
\providecommand{\BIBentryALTinterwordstretchfactor}{4}
\providecommand{\BIBentryALTinterwordspacing}{\spaceskip=\fontdimen2\font plus
\BIBentryALTinterwordstretchfactor\fontdimen3\font minus \fontdimen4\font\relax}
\providecommand{\BIBforeignlanguage}[2]{{%
\expandafter\ifx\csname l@#1\endcsname\relax
\typeout{** WARNING: IEEEtran.bst: No hyphenation pattern has been}%
\typeout{** loaded for the language `#1'. Using the pattern for}%
\typeout{** the default language instead.}%
\else
\language=\csname l@#1\endcsname
\fi
#2}}
\providecommand{\BIBdecl}{\relax}
\BIBdecl

\bibitem{RISeg}
H.~H. Qian, Y.~Lu, K.~Ren, G.~Wang, N.~Khargonkar, Y.~Xiang, and K.~Hang, ``Riseg: Robot interactive object segmentation via body frame-invariant features,'' in \emph{IEEE International Conference on Robotics and Automation (ICRA)}.\hskip 1em plus 0.5em minus 0.4em\relax IEEE, 2024.

\bibitem{Back2022Amodal}
S.~Back, J.~Lee, T.~Kim, S.~Noh, R.~Kang, S.~Bak, and K.~Lee, ``Unseen object amodal instance segmentation via hierarchical occlusion modeling,'' in \emph{IEEE International Conference on Robotics and Automation (ICRA)}.\hskip 1em plus 0.5em minus 0.4em\relax IEEE, 2022, pp. 5085--5092.

\bibitem{Xie2020UOIS}
C.~Xie, Y.~Xiang, A.~Mousavian, and D.~Fox, ``Unseen object instance segmentation for robotic environments,'' \emph{IEEE Transactions on Robotics}, vol.~37, no.~5, pp. 1343--1359, 2021.

\bibitem{Xiang2021UOIS}
Y.~Xiang, C.~Xie, A.~Mousavian, and D.~Fox, ``Learning rgb-d feature embeddings for unseen object instance segmentation,'' in \emph{Conference on Robot Learning}.\hskip 1em plus 0.5em minus 0.4em\relax PMLR, 2021, pp. 461--470.

\bibitem{Danielczuk2018}
M.~Danielczuk, M.~Matl, S.~Gupta, A.~Li, A.~Lee, J.~Mahler, and K.~Goldberg, ``Segmenting unknown 3d objects from real depth images using mask r-cnn trained on synthetic point clouds,'' \emph{IEEE International Conference on Robotics and Automation (ICRA)}, 2019.

\bibitem{MSMFormer}
Y.~Lu, Y.~Chen, N.~Ruozzi, and Y.~Xiang, ``Mean shift mask transformer for unseen object instance segmentation,'' in \emph{IEEE International Conference on Robotics and Automation (ICRA)}.\hskip 1em plus 0.5em minus 0.4em\relax IEEE, 2024.

\bibitem{Brock2009}
J.~Kenney, T.~Buckley, and O.~Brock, ``Interactive segmentation for manipulation in unstructured environments,'' in \emph{IEEE International Conference on Robotics and Automation (ICRA)}.\hskip 1em plus 0.5em minus 0.4em\relax IEEE, 2009, pp. 1377--1382.

\bibitem{ModernRobotics}
K.~M. Lynch and F.~C. Park, \emph{Modern robotics}.\hskip 1em plus 0.5em minus 0.4em\relax Cambridge University Press, 2017.

\bibitem{MARTINEZContPercept}
L.~Martínez, J.~R. del Solar, L.~Sun, J.~P. Siebert, and G.~Aragon-Camarasa, ``Continuous perception for deformable objects understanding,'' \emph{Robotics and Autonomous Systems}, vol. 118, pp. 220--230, 2019.

\bibitem{kirillov2023segmentanything}
A.~Kirillov, E.~Mintun, N.~Ravi, H.~Mao, C.~Rolland, L.~Gustafson, T.~Xiao, S.~Whitehead, A.~C. Berg, W.-Y. Lo \emph{et~al.}, ``Segment anything,'' in \emph{International Conference on Computer Vision}.\hskip 1em plus 0.5em minus 0.4em\relax IEEE, 2023, pp. 4015--4026.

\bibitem{Felzenszwalb2004Graph}
P.~F. Felzenszwalb and D.~P. Huttenlocher, ``Efficient graph-based image segmentation,'' \emph{International Journal of Computer Vision}, vol.~59, pp. 167--181, 2004.

\bibitem{Pham2017SceneCut}
T.~T. Pham, T.-T. Do, N.~S{\"u}nderhauf, and I.~Reid, ``Scenecut: Joint geometric and object segmentation for indoor scenes,'' in \emph{IEEE International Conference on Robotics and Automation (ICRA)}.\hskip 1em plus 0.5em minus 0.4em\relax IEEE, 2018, pp. 3213--3220.

\bibitem{Stein2014Convexity}
S.~Christoph~Stein, M.~Schoeler, J.~Papon, and F.~Worgotter, ``Object partitioning using local convexity,'' in \emph{IEEE Conference on Computer Vision and Pattern Recognition}, 2014, pp. 304--311.

\bibitem{forsyth2003computer}
D.~A. Forsyth and J.~Ponce, \emph{Computer vision: a modern approach}.\hskip 1em plus 0.5em minus 0.4em\relax prentice hall professional technical reference, 2002.

\bibitem{Trevor2013EfficientOP}
A.~J. Trevor, S.~Gedikli, R.~B. Rusu, and H.~I. Christensen, ``Efficient organized point cloud segmentation with connected components,'' \emph{Semantic Perception Mapping and Exploration (SPME)}, vol.~10, no.~6, pp. 251--257, 2013.

\bibitem{Xie2019Both}
C.~Xie, Y.~Xiang, A.~Mousavian, and D.~Fox, ``The best of both modes: Separately leveraging rgb and depth for unseen object instance segmentation,'' in \emph{Conference on Robot Learning}.\hskip 1em plus 0.5em minus 0.4em\relax PMLR, 2020, pp. 1369--1378.

\bibitem{Danielczuk2019Data}
M.~Danielczuk, M.~Matl, S.~Gupta, A.~Li, A.~Lee, J.~Mahler, and K.~Goldberg, ``Segmenting unknown 3d objects from real depth images using mask r-cnn trained on synthetic data,'' in \emph{IEEE International Conference on Robotics and Automation (ICRA)}.\hskip 1em plus 0.5em minus 0.4em\relax IEEE, 2019, pp. 7283--7290.

\bibitem{Shao2018ClusterNetIS}
L.~Shao, Y.~Tian, and J.~Bohg, ``Clusternet: 3d instance segmentation in rgb-d images,'' \emph{arXiv preprint arXiv:1807.08894}, 2018.

\bibitem{Zhang2023UOIS}
L.~Zhang, S.~Zhang, X.~Yang, H.~Qiao, and Z.~Liu, ``Unseen object instance segmentation with fully test-time rgb-d embeddings adaptation,'' in \emph{IEEE International Conference on Robotics and Automation (ICRA)}.\hskip 1em plus 0.5em minus 0.4em\relax IEEE, 2023, pp. 4945--4952.

\bibitem{Balloch2018Synthetic}
J.~C. Balloch, V.~Agrawal, I.~Essa, and S.~Chernova, ``Unbiasing semantic segmentation for robot perception using synthetic data feature transfer,'' \emph{arXiv preprint arXiv:1809.03676}, 2018.

\bibitem{Bohg2016}
J.~Bohg, K.~Hausman, B.~Sankaran, O.~Brock, D.~Kragic, S.~Schaal, and G.~S. Sukhatme, ``Interactive perception: Leveraging action in perception and perception in action,'' \emph{IEEE Transactions on Robotics}, vol.~33, no.~6, pp. 1273--1291, 2017.

\bibitem{zappella2008motion}
C.~Julia, A.~Sappa, F.~Lumbreras, J.~Serrat, and A.~L{\'o}pez, ``Motion segmentation from feature trajectories with missing data,'' in \emph{Pattern Recognition and Image Analysis}.\hskip 1em plus 0.5em minus 0.4em\relax Springer, 2007, pp. 483--490.

\bibitem{Costeira1998AMF}
J.~P. Costeira and T.~Kanade, ``A multibody factorization method for independently moving objects,'' \emph{International Journal of Computer Vision}, vol.~29, pp. 159--179, 1998.

\bibitem{Goh2007}
A.~Goh and R.~Vidal, ``Segmenting motions of different types by unsupervised manifold clustering,'' in \emph{IEEE Conference on Computer Vision and Pattern Recognition}.\hskip 1em plus 0.5em minus 0.4em\relax IEEE, 2007, pp. 1--6.

\bibitem{Arsenio}
A.~Arsenio, P.~Fitzpatrick, C.~C. Kemp, and G.~Metta, ``The whole world in your hand: Active and interactive segmentation,'' in \emph{Proceedings of the Third International Workshop on Epigenetic Robotics}, 2003, pp. 49--56.

\bibitem{metta2003early}
G.~Metta and P.~Fitzpatrick, ``Early integration of vision and manipulation,'' \emph{Adaptive behavior}, vol.~11, no.~2, pp. 109--128, 2003.

\bibitem{Zeng2017MultiView}
A.~Zeng, K.-T. Yu, S.~Song, D.~Suo, E.~Walker, A.~Rodriguez, and J.~Xiao, ``Multi-view self-supervised deep learning for 6d pose estimation in the amazon picking challenge,'' in \emph{IEEE International Conference on Robotics and Automation (ICRA)}.\hskip 1em plus 0.5em minus 0.4em\relax IEEE, 2017, pp. 1386--1383.

\bibitem{Mitash2017MultiView}
C.~Mitash, K.~E. Bekris, and A.~Boularias, ``A self-supervised learning system for object detection using physics simulation and multi-view pose estimation,'' in \emph{IEEE International Conference on Intelligent Robots and Systems (IROS)}.\hskip 1em plus 0.5em minus 0.4em\relax IEEE, 2017, pp. 545--551.

\bibitem{lu2023selfsupervised}
Y.~Lu, N.~Khargonkar, Z.~Xu, C.~Averill, K.~Palanisamy, K.~Hang, Y.~Guo, N.~Ruozzi, and Y.~Xiang, ``Self-supervised unseen object instance segmentation via long-term robot interaction,'' in \emph{Robotics: Science and Systems}, 2023.

\bibitem{fischler1981ransac}
M.~FISCHLER~AND, ``Random sample consensus: a paradigm for model fitting with applications to image analysis and automated cartography,'' \emph{Commun. ACM}, vol.~24, no.~6, pp. 381--395, 1981.

\bibitem{macqueen1967kmeans}
J.~MacQueen, ``Some methods for classification and analysis of multivariate observations,'' in \emph{Proceedings of the Fifth Berkeley Symposium on Mathematical Statistics and Probability, Volume 1: Statistics}, vol.~5.\hskip 1em plus 0.5em minus 0.4em\relax University of California press, 1967, pp. 281--298.

\bibitem{kmedoids}
L.~Kaufmann and P.~Rousseeuw, ``Clustering by means of medoids,'' \emph{Data Analysis based on the L1-Norm and Related Methods}, pp. 405--416, 01 1987.

\bibitem{meanshift}
D.~Comaniciu and P.~Meer, ``Mean shift: a robust approach toward feature space analysis,'' \emph{IEEE Transactions on Pattern Analysis and Machine Intelligence}, vol.~24, no.~5, pp. 603--619, 2002.

\bibitem{kucuk2006forwardkinematics}
S.~Kucuk and Z.~Bingul, \emph{Robot kinematics: Forward and inverse kinematics}.\hskip 1em plus 0.5em minus 0.4em\relax INTECH Open Access Publisher London, UK, 2006.

\bibitem{Mahalanobis}
R.~H. Riffenburgh, \emph{Statistics in medicine}.\hskip 1em plus 0.5em minus 0.4em\relax Academic press, 2012.

\bibitem{van2000graph}
S.~Van~Dongen, ``Graph clustering by flow simulation,'' \emph{PhD thesis, University of Utrecht}, 2000.

\bibitem{FrankaEmika}
S.~Haddadin, S.~Parusel, L.~Johannsmeier, S.~Golz, S.~Gabl, F.~Walch, M.~Sabaghian, C.~J{\"a}hne, L.~Hausperger, and S.~Haddadin, ``The franka emika robot: A reference platform for robotics research and education,'' \emph{IEEE Robotics and Automation Magazine}, vol.~29, no.~2, pp. 46--64, 2022.

\bibitem{D415}
M.~Carfagni, R.~Furferi, L.~Governi, C.~Santarelli, M.~Servi, F.~Uccheddu, and Y.~Volpe, ``Metrological and critical characterization of the intel d415 stereo depth camera,'' \emph{Sensors}, vol.~19, no.~3, p. 489, 2019.

\bibitem{RAFT}
Z.~Teed and J.~Deng, ``Raft: Recurrent all-pairs field transforms for optical flow,'' in \emph{European Conference on Computer Vision}, 2020, pp. 402--419.

\end{thebibliography}
